\documentclass{article}

\PassOptionsToPackage{numbers, compress}{natbib}
\usepackage[final]{neurips_2024}
\usepackage{natbib}

\usepackage[utf8]{inputenc} 
\usepackage[T1]{fontenc}    
\usepackage{hyperref}       
\usepackage{url}            
\usepackage{booktabs}       
\usepackage{amsfonts}       
\usepackage{nicefrac}       
\usepackage{microtype}      
\usepackage{xcolor}         
\usepackage{makecell}
\usepackage{graphicx}
\usepackage{amsmath}
\usepackage{amssymb}
\usepackage{booktabs}
\usepackage{multirow}
\usepackage{enumitem}

\usepackage{amsmath,amsfonts,bm}

\def\eqref#1{equation~\ref{#1}}

\def\1{\bm{1}}

\DeclareMathAlphabet{\mathsfit}{\encodingdefault}{\sfdefault}{m}{sl}
\SetMathAlphabet{\mathsfit}{bold}{\encodingdefault}{\sfdefault}{bx}{n}

\usepackage[capitalize]{cleveref}
\crefname{section}{Sec.}{Secs.}
\Crefname{section}{Section}{Sections}
\Crefname{table}{Table}{Tables}
\crefname{table}{Tab.}{Tabs.}

\newcommand{\ie}{\textit{i}.\textit{e}.}
\newcommand{\eg}{\textit{e}.\textit{g}.}

\title{Multi-times Monte Carlo Rendering for Inter-reflection Reconstruction}

\author{Tengjie Zhu \thanks{Equal contribution.}  ~\quad  Zhuo Chen \textsuperscript{\rm *}  ~\quad Jingnan Gao ~\quad  Yichao Yan\thanks{\hspace{0em}Corresponding author.} ~\quad Xiaokang Yang \\\\\
MoE Key Lab of Artificial Intelligence, AI Institute, Shanghai Jiao Tong University \\
\texttt{\{zhutengjie, ningci5252, gjn0310, yanyichao, xkyang\}@sjtu.edu.cn}} 

\begin{document}

\maketitle

\begin{abstract}
Inverse rendering methods have achieved remarkable performance in reconstructing high-fidelity 3D objects with disentangled geometries, materials, and environmental light. However, they still face huge challenges in reflective surface reconstruction. Although recent methods model the light trace to learn specularity, the ignorance of indirect illumination makes it hard to handle inter-reflections among multiple smooth objects. 
In this work, we propose Ref-MC$^2$ that introduces the multi-times Monte Carlo sampling which comprehensively computes the environmental illumination and meanwhile considers the reflective light from object surfaces.
To address the computation challenge as the times of Monte Carlo sampling grow, we propose a specularity-adaptive sampling strategy, significantly reducing the computational complexity. Besides the computational resource, higher geometry accuracy is also required because geometric errors accumulate multiple times. Therefore, we further introduce a reflection-aware surface model to initialize the geometry and refine it during inverse rendering.
We construct a challenging dataset containing scenes with multiple objects and inter-reflections.
Experiments show that our method outperforms other inverse rendering methods on various object groups. We also show downstream applications, \eg, relighting and material editing, to illustrate the disentanglement ability of our method. Our project page: \url{https://zhutengjie.github.io/Ref-MC2/}.

\end{abstract}


\section{Introduction}
\label{sec:intro}

Neural Radiance Fields (NeRF)~\cite{nerf} and 3DGS~\cite{3dgs} have demonstrated their excellent performance on novel view synthesis. However, 
it is difficult to directly apply their reconstructed 3D model to the current industrial pipeline, leading to the lack of flexibility in many downstream applications, \eg, relighting and material editing. 
To better cooperate with mature techniques, inverse rendering bases the physical rendering~\cite{mcauley2012practical} and utilizes the neural network to learn disentangled materials that can be seamlessly plunged into the industrial pipeline for further manipulation. 

Previous methods~\cite{munkberg2022extracting, hasselgren2022shape, wu2023nefii, DBLP:conf/cvpr/ZhangSHFJZ22} have explored how to disentangle geometry, diffuse, roughness, metalness, and environmental light from multi-view images, but they still face challenges in shading and reflective objects.
Specifically, Nvdiffrec~\cite{munkberg2022extracting} proposes a differentiable pipeline that enables gradient-based optimization on both meshes and volumetric textures.
These methods ignore the shadow when modeling the illumination, resulting in failed disentanglement for diffuse and shading appearances.
To model more realistic shadings, Nvdiffrecmc~\cite{hasselgren2022shape} further incorporates ray tracing and Monte Carlo sampling~\cite{morokoff1995quasi} into inverse rendering, significantly improving the decomposition of shape, materials, and lighting.
However, it ignores the indirect illuminations during path tracing and treats rays attacking object surfaces as sheltered illumination. 
As a consequence, methods like Nvdiffrecmc increase the ambiguity of geometry reconstruction and undermine material learning.
Unfortunately, the scene with multiple inter-reflections is common in the real world. The failure in these scenes hinders the wider applications of these methods.
Recent methods~\cite{DBLP:conf/iccv/LiangC0CPV23,verbin2022ref,mai2023neural} take inter-reflections into account and can reconstruct the reflective objects well. However, the implicit representations of these methods for materials and renders sacrifice their scalability to downstream tasks.
Recently, Nefii~\cite{wu2023nefii} further incorporated the implicit neural radiance to estimate ray tracing, alleviating the ambiguity between materials and indirect illuminations. However, this leads to a huge computational consumption for path tracing. Neural Microfacet Fields~\cite{mai2023neural} employs two-bounce sampling for indirect light modeling, but its geometry field limits flexibility for downstream tasks.


In this paper, we proposed a full inverse rendering method, Ref-MC$^2$, which considers inter-reflections during ray tracing to improve the decomposition of explicit materials and environmental lighting. 
The core of our method is to use \textbf{Multi-times Monte Carlo integration}~\cite{morokoff1995quasi} and \textbf{BRDF}~\cite{mcauley2012practical} rendering to approximate the indirect illumination at multiple reflection points along the light propagation path.
Although our method takes advantage of hardware-accelerated ray tracing to model the indirect illumination,
we still face two major challenges brought by the multi-times Monte Carlo sampling.
1) \textbf{Efficiency}: the explosive computational growth from multiple times is too heavy for the hardware algorithm only.
2) \textbf{Geometry}: the geometry quality greatly affects the calculation of indirect light and the decomposition of the materials, because the error will accumulate over and over again as the times of Monte Carlo sampling increase.

To address these challenges, we correspondingly propose two strategies. 
1) For efficiency, 
we flatten the multi-times sampling into sequential single-time sampling.
In a Lambert model~\cite{oren1994generalization}, the diffuse light is independent of the direction of reflection and thus can be presented as a map to query at any time.
When we trace the indirect light from an object, instead of recalculating the diffuse component, we can take the value directly from the diffuse map. It can be optimized through self-supervision.
Therefore, we only need to sample the specular component within a small lobe along around the reflective direction. 
2) For geometry,
we refer to the SDF-based methods~\cite{unisdf, DBLP:conf/iccv/LiangC0CPV23, ge2023ref} and replace the common positional encoding with the Sphere Gaussian encoding to get an accurate initial geometry for reflective objects.
We use this geometry to initialize Flxicubes ~\cite{shen2023flexible} that optimizes surface meshes based on the gradient, and further fine-tunes the Flexicubes in a differentiable pipeline. 
To evaluate our framework, we construct a dataset containing difficult scenes which contain strong inter-reflections between multiple surfaces. Extensive experiments demonstrate that our framework can successfully decompose indirect illumination and materials.
In summary, our contributions are:
\begin{itemize}[leftmargin=2.5em]
\setlength\itemsep{0em}
\item We propose a full inverse rendering method that employs multi-times Monte Carlo sampling to correctly decompose indirect illumination and materials.
\item We reduce the computational consumption when tracing indirect illumination by self-supervising the diffuse map based on the Lambert model. 
\item We refine the SDF-based architecture with Spherical Gaussian encoding to obtain a high-quality initial geometry which further releases the accumulated error during multi-times sampling. 
\item We construct a dataset to evaluate the performance on indirect illumination.

\end{itemize}

\section{Related Work}
\label{sec:related}

\subsection{Implicit Neural Representations}
Neural implicit representations~\cite{Martin-BruallaR21, nerf,DBLP:conf/iccv/ParkSBBGSM21,DBLP:journals/corr/abs-2010-07492,wang2021nerfmm,DBLP:journals/tog/ReiserSVSMGBH23, DBLP:conf/iccv/HuWMY0LM23,DBLP:conf/eccv/ChenXGYS22,DBLP:conf/iccv/ChenLSCYYX23,DBLP:conf/iccv/Zhang0YYJWY023,DBLP:conf/siggrapha/ShuYMWM23,vmesh} have achieved impressive success in many computer vision and computer graphics tasks. 
These methods use neural radiance fields to capture color and volume density, generating photo-realistic novel views through volume rendering~\cite{kajiya1984ray}. However, the unconstrained volumetric representation of the
original NeRF method leads to low-quality geometry.
Recent 3D Gaussian Splatting (3DGS)~\cite{3dgs, 2dgs,gshader,scaffoldgs,Yu2023MipSplatting} has gained popularity in novel view synthesis. Different from NeRF, it is an explicit representation that involves the optimization of multiple Gaussians to reconstruct 3D objects. 3DGS learns color and density in a volumetric point cloud, but it also fails to produce accurate geometry due to its discrete representation.
Besides, these methods are all entangled learning, integrating all inherent materials and the environment map into the appearance.
This limits the downstream applications in the current industrial pipeline.
To produce a high-quality geometry, follow-up works ~\cite{yariv2021volume, wang2021neus, yariv2020multiview, oechsle2021unisurf} use a function to associate signed distance field (SDF) and volume density. The surface mesh can be extracted from neural implicit surfaces by Marching Cubes~\cite{lorensen1998marching}, and this 3D asset can be further applied in other applications.
However, they usually perform badly in reconstructing the reflective objects due to the ambiguity of reflective appearance.



\subsection{Neural Inverse Rendering}
Although neural implicit surfaces have achieved impressive performance in geometry reconstructing and novel views 
synthesis, they do not obtain the fundamental materials of PBR which limits their flexibility in downstream tasks.
Neural inverse rendering methods~\cite{DBLP:conf/iccv/LiangC0CPV23,DBLP:journals/tog/WuXZBHTX22,DBLP:conf/cvpr/GuoKBHZ22,DBLP:conf/nips/BossJBLBL21,DBLP:conf/cvpr/ZhangLWBS21,DBLP:conf/cvpr/ZhangSHFJZ22,DBLP:conf/cvpr/JinLXZHBZX023,DBLP:conf/nips/TangZWS23,DBLP:conf/ijcai/Lv0CLS23} introduce the physical rendering equation to estimate the disentangled diffuse and specular component from RGB images. They approximate the rendering equation based on neural networks or basis functions, \eg, Spherical Gaussians~\cite{DBLP:journals/tog/WangRGSG09,DBLP:journals/tog/XuSDZWH13,bakedsdf,DBLP:conf/cvpr/ZhangLWBS21} and Spherical Harmonics~\cite{yu2022plenoxels,DBLP:journals/pami/BasriJ03,DBLP:journals/tog/SloanKS02,DBLP:conf/iccv/YuLT0NK21}.
Nvdiffrec~\cite{munkberg2022extracting} introduces the \textbf{full inverse rendering} that estimates shape, materials, and environmental light into gradient-based optimization. However, it does not consider the shadows, leading to the entanglement of materials.
Recent works~\cite{ge2023ref,unisdf} extend SDF-based architectures with an additional appearance branch to model the reflections on the object.  RefNeuS~\cite{ge2023ref} introduces a reparametrization method to distinguish the reflective appearance from the diffuse appearance by a direction-relative process, but it fails to faithfully reconstruct non-reflective objects.
UniSDF~\cite{unisdf} proposes to use a weight-MLP to balance the reflective and non-reflective branches for different objects. 
However, these methods still face challenges in scenes with complex inter-reflections.
Nvdiffrecmc~\cite{hasselgren2022shape} extends Nvdiffrec with Monte Carlo sampling to trace the light path but still ignores indirect illumination between objects.
Further methods~\cite{DBLP:conf/iccv/LiangC0CPV23,liu2023nero,wu2023nefii} consider indirect illumination in their design.
ENVIDR~\cite{DBLP:conf/iccv/LiangC0CPV23} employs a neural renderer to learn the physical light interaction, without explicitly formulating the rendering equation. 
NeRO~\cite{liu2023nero} applies the split-sum approximation to approximate the shading effects of both direct and indirect lights.
Nefii~\cite{wu2023nefii} introduces ray tracing to the radiance field to model indirect illuminations. Neural Microfacet Fields~\cite{mai2023neural} employs two-bounce sampling to accurately calculate indirect illumination. However, NMF represents geometry by a density field, which limits the scalability and flexibility for downstream tasks. In addition, these aforementioned methods do not fully disentangle the materials from RGB images, but decompose the appearance into reflective and diffuse color. It is hard to apply to the current industrial pipeline directly.eq: rendering equation
In contrast, our Ref-MC$^2$ is a full inverse rendering method that also considers indirect illumination.


\section{Method}
\subsection{Preliminaries}
\label{prelim}
The rendering equation~\cite{kajiya1986rendering} is commonly used to compute the outgoing radiance $L_o$($\mathbf{p}$, $\boldsymbol{\omega}_o$) from the point $\mathbf{p}$ in outgoing direction $\boldsymbol{\omega}_o$:
\begin{equation}
L_o\left(\mathbf{p}, \boldsymbol{\omega}_o\right)=\int_{\Omega} L_i\left(\mathbf{p}, \boldsymbol{\omega}_i\right) f\left(\mathbf{p}, \boldsymbol{\omega}_i, \boldsymbol{\omega}_o\right)\left(\mathbf{n}\cdot \boldsymbol{\omega}_i\right) d \boldsymbol{\omega}_i,
\label{eq: rendering equation}
\end{equation}
where $L_i$($\mathbf{p}$, $\boldsymbol{\omega}_i$) is the incoming radiance into $\mathbf{p}$ from the direction $\boldsymbol{\omega}_i$, $\mathbf{n}$ is the normal of the point $\mathbf{p}$, $\Omega$ is the hemisphere of directions above $\mathbf{p}$, $f(\mathbf{p}, \boldsymbol{\omega}_i, \boldsymbol{\omega}_o)$ is the BSDF evaluated for $\boldsymbol{\omega}_i$ and the current incoming direction $\boldsymbol{\omega}_i$. 
The GGX~\cite{mcauley2012practical} physics-based BSDF function is proposed to decompose the function into several physical terms.
The function is described as:
\begin{equation}
f\left(\mathbf{p}, \boldsymbol{\omega}_i, \boldsymbol{\omega}_o\right) = f_d + f_s = f_d + \frac{D F G}{4 \left(\mathbf{n}\cdot \boldsymbol{\omega}_i\right) \left(\mathbf{n}\cdot \boldsymbol{\omega}_o\right)},
\label{eq: BSDF}
\end{equation}
where $f_d$ is the diffuse term and $f_s$ is the specular term. 
$D$, $F$, and $G$ are the microfacet distribution function, the Fresnel reflection coefficient, and the geometric attenuation, respectively.

Previous works~\cite{munkberg2022extracting} use the split sum function~\cite{mcauley2013physically} to approximate the rendering equation, but it inevitably omits the shadow and indirect illumination. 
In contrast, 
Monte Carlo integration~\cite{morokoff1995quasi} is a simple but unbiased estimation method that comprehensively considers all the physical terms for the outgoing radiance. The Monte Carlo integration rendering equation is:
\begin{equation}
L_o\left(\mathbf{p},\boldsymbol{\omega}_o\right) \approx \frac{1}{N} \sum_{i=1}^N \frac{L_i\left(\mathbf{p},\boldsymbol{\omega}_i\right) f\left(\mathbf{p}, 
\boldsymbol{\omega}_i, \boldsymbol{\omega}_o\right)\left(\boldsymbol{\omega}_i \cdot \mathbf{n}\right)}{p\left(\boldsymbol{\omega}_i\right)},
\label{eq: rendering equation2}
\end{equation}
where $\boldsymbol{\omega}_i$ is the $i$th sample drawn from density $p$.
As the number of samples grows, the estimation variance reduces, but the computation increases.
Multiple importance sampling~\cite{veach1995optimally} (MIS) is proposed to inhibit the computational consumption.
When a function can be expressed as a multiplication of $n$ functions, it draws $n_i$ samples $\boldsymbol{\omega}_{i, j}$ from $n$ sampling distributions $p_i$ in turn.
The MIS Monte Carlo estimator for the rendering equation is:
\begin{equation}
L_o\left(\mathbf{p},\boldsymbol{\omega}_o\right) =\sum_{i=1}^n \frac{1}{n_i} \sum_{j=1}^{n_i} W_i\frac{F_i\left(\mathbf{p},\boldsymbol{\omega}_o,\boldsymbol{\omega}_{i, j}\right)}{p_i},
\label{eq: MIS}
\end{equation}
where $p_i$ $\propto$ the $i$-th multiplication function of the integrated function of the rendering equation and $W_i$ is the the balance heuristic weighting function.

\begin{figure}[t]
  \centering
  \includegraphics[width=\linewidth]{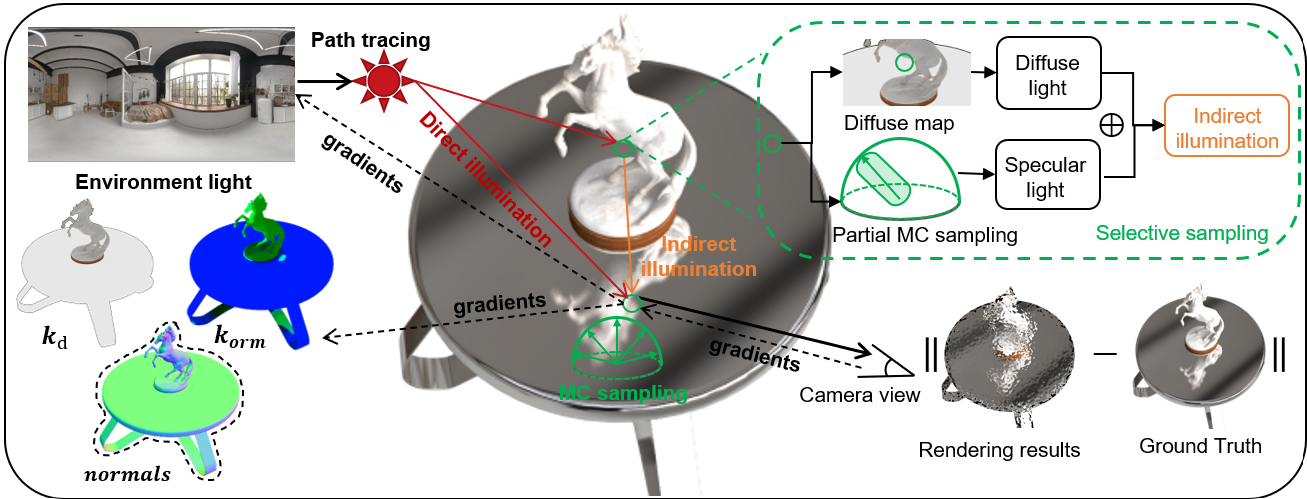}
   \caption{We perform Monte Carlo sampling at the viewpoint. When the sampling ray from the point is not blocked, it is the direct illumination from environmental lighting. When the sampling ray hits an object, we divide this indirect illumination from the object into diffuse light and specular light.
   We sample the diffuse light from a diffuse map that is optimized through self-supervision. For specular light, we only need to partially trace the rays in a small specular lobe along the reflective direction. The gradients are backward along the tracing path, and are passed to optimize $\boldsymbol{k}_d, \boldsymbol{k}_{orm}$, normals, and environment maps.}
   \label{fig:method_1}
\end{figure}

\subsection{Multi-times Monte Carlo Sampling}
\label{sec:Multi-times}

As discussed in ~\cref{prelim}, the split sum approximation ignores the shadows and the indirect illumination. 
Nvdiffrecmc notes this and accounts for shadows using the Monte Carlo integration method. 
However, they give up continuously tracing the light rays attacking the surface of objects and treat them as zero illumination.
It leads to bad performance in scenes with inter-reflections.
Therefore, we propose the Ref-MC$^2$ method to trace the light rays continuously. 
When taking into account indirect illumination, based on ~\cite{kajiya1986rendering}, the rendering equation can be expressed in this version:
\begin{equation}
L_o\left(\mathbf{p}, \boldsymbol{\omega}_o\right)=\int_{\Omega} L_i\left(r\left(\mathbf{p},\boldsymbol{\omega}_i\right) ,-\boldsymbol{\omega}_i\right) f\left(\mathbf{p}, \boldsymbol{\omega}_i, \boldsymbol{\omega}_o\right)\left(\mathbf{n}\cdot \boldsymbol{\omega}_i\right) d \boldsymbol{\omega}_i.
\label{eq: rendering}
\end{equation}

In this equation, $r\left(\mathbf{p},\boldsymbol{\omega}_i\right)$ represents the location of a surface point on the object surface hit by a ray cast from $\mathbf{p}$ in direction $\boldsymbol{\omega}_i $ for the first time. The corresponding Monte Carlo integration that considers the indirect illumination can then be expressed as:

\begin{equation}
L_o\left(\mathbf{p},\boldsymbol{\omega}_o\right) \approx \frac{1}{N} \sum_{i=1}^N \frac{L_i\left(r\left(\mathbf{p},\boldsymbol{\omega}_i\right) ,-\boldsymbol{\omega}_i\right) f\left(\mathbf{p}, 
\boldsymbol{\omega}_i, \boldsymbol{\omega}_o\right)\left(\boldsymbol{\omega}_i \cdot \mathbf{n}\right)}{p\left(\boldsymbol{\omega}_i\right)}.
\label{eq: rendering equation3}
\end{equation}
It can be interpreted as that when the sampling ray is not blocked, it can be seen as direct illumination from environmental lighting.
When the surface point of an object blocks the sampling ray, it needs to be continuously traced from this surface point. This continuous tracing is an iterated process of Monte Carlo integration.
It is noteworthy that when continuous sampling at the blocking surface point it is unnecessary to consider the indirect illumination like calculating the outgoing radiance at initial point $\mathbf{p}$.
This is because the energy of light gradually degrades and the impact of light after twice reflections is negligible compared to the explosively increased computational load. 
However, the additional computational load of ray tracing at a depth of two is still huge. To reduce our computational load, we further propose to approximate the diffuse light and transfer the computations. 

In Disney PBR equation~\cite{mcauley2012practical}, the diffuse term $f_d$ is:
\begin{equation}
f_d=\frac{c_{\text{diff}}}{\pi}\left(1+\left(F_{D 90}-1\right)\left(1-\left(\mathbf{n} \cdot \boldsymbol{\omega}_i\right)\right)^5\right)\left(1+\left(F_{D 90}-1\right)\left(1-\left(\mathbf{n} \cdot \boldsymbol{\omega}_o\right)\right)^5\right),
\label{eq: diffuse}
\end{equation}

\begin{equation}
F_{D 90}=0.5+2r\cos ^2 \theta_d,
\label{FD}
\end{equation}
where $c_{\text{diff}}$ is the diffuse albedo of the material. In this equation, $f$ is related to the direction of the incoming radiance, which can yield more realistic results. 
However, using this equation to calculate the diffuse light for indirect illumination creates unnecessary extra computation.
Because the term $f_d$ is related to the normal $n$ and the direction of rays ${\omega}_o$, we cannot get the diffuse map before deep ray tracing.
The following term is to introduce the roughness for diffuse light to avoid too dark edges in extremely low grazing angles. 
However, using this equation to compute the diffuse light for indirect illumination introduces unnecessary extra computation. To reduce the burden, we can follow H and approximate it using Lambertian diffuse lighting.:
\begin{equation}
f_{d}^{\text{ind}}=\frac{c_{\text{diff}}}{\pi}.
\label{eq: Lambert}
\end{equation}

Where $f_{d}^{\text{ind}}$ is the diffuse light of the indirect illumination. In this approximation, $f_{d}^{\text{ind}}$ is independent on the incoming direction $\omega_i$ and the
outgoing direction $\omega_o$. Based on this, the diffuse part of indirect light can be simplified as:
\begin{equation}
L_{\text{diff}}^{\text{ind}}\left(\mathbf{p} \right)=\frac{c_{\text{diff}}}{\pi} \int_{\Omega} L_i\left(\mathbf{p}, \boldsymbol{\omega}_i\right)\left(\mathbf{n}\cdot \boldsymbol{\omega}_i\right) d \boldsymbol{\omega}_i,
\label{equation}
\end{equation}
where $L_{\text{diff}}^{\text{ind}}$ is the diffuse part of the indirect illumination. It no longer relates to the direction of the sampling ray, so we can present it via an MLP $M$. 
Compared to the diffuse part, the specular part is strongly related to the ray direction. It is a disadvantage that we cannot approximate the specular part like the diffuse part, but it is also an advantage that we only need to sample minor rays in a small specular lobe along the reflective directions. 
Therefore, we significantly reduce the computations of both the diffuse part and the specular part during the multi-times Monte Carlo Sampling.

\begin{figure}[t]
  \centering
  \includegraphics[width=\linewidth]{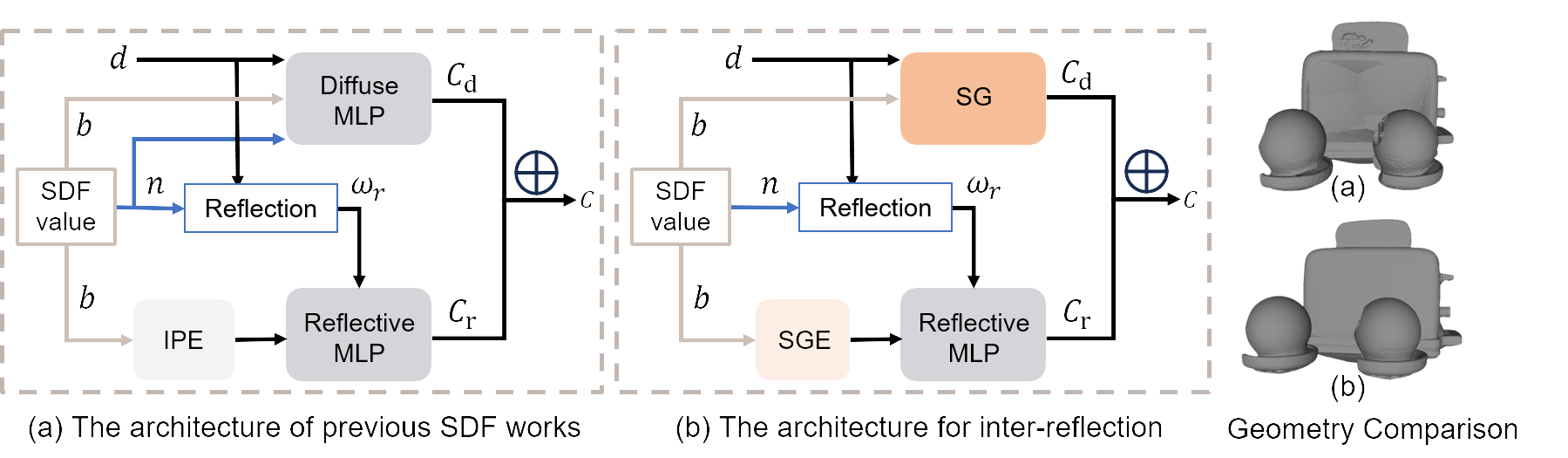}
   \caption{\textbf{Differences} between previous SDF architectures and the architecture for inter-reflections.}
   \label{fig:geo_pipeline}
\end{figure}

\subsection{Geometry Initialization}
\label{sec:geometry}
As seen in the equation \ref{eq: rendering}, our indirect lighting highly depends on the geometry. 
However, the strong ambiguity of reflections makes it hard to directly learn high-quality disentangled shapes.
Therefore, we flatten the joint learning process of both geometry and materials into separate learning of them and thus reduce the number of physical terms to be disentangled in each learning stage. 

We first utilize SDF-based architectures~\cite{yariv2021volume, ge2023ref, liu2023nero} to learn an initial geometry. The two branches of the diffuse and reflective networks well disambiguate the appearance with reflections and empower the SDF network to produce a high-quality geometry, as shown in~\cref{fig:geo_pipeline}~(a).
However, this architecture encounters challenges of indirect illumination due to the expressive capacity of Integrated Positional Encoding (IPE). It performs well in general scenes but shows limitations in representing inter-reflections.
Due to the capacity of Spheical Gaussians (SG)~\cite{yariv2023bakedsdf, reiser2024binary} to represent radiance directions, we introduce SG encoding instead of IPE to enhance the expressive capacity of reflective MLP, as shown in~\cref{fig:geo_pipeline}~(b).
Besides, we directly parameterize diffuse appearance as SG coefficients which is suitable for objects with multiple reflective surfaces. The geometry comparison is shown in~\cref{fig:geo_pipeline}.


The learned SDF can be then converted into a surface mesh using the Marching Cubes~\cite{lorensen1998marching}, which supports further tuning in the following inverse rendering pipeline.
Different from Nvdiffrecmc which adopts DMTET for differentiable mesh optimization, we introduce Flexicubes~\cite{shen2023flexible} into our inverse rendering pipeline.
Flexicubes use the SDF, weight, and the deformation for vertexes of the grid cells to extract the surface mesh using the DMC~\cite{nielson2004dual}. 
The mesh is converted into a differentiable representation that can be optimized based on gradients. 


\subsection{Training Objectives}
The main objective is to minimize the photometric loss between rendered images and ground truth:
\begin{equation}
    \mathcal{L}_{rgb} = ||C - C_{gt}||^2,
\end{equation}
where $C$ is the rendering result, and $C_{gt}$ is the corresponding ground truth. Following previous work~\cite{zhang2021nerfactor, munkberg2022extracting, hasselgren2022shape}, we adopt a smoothness loss for diffuse and material as regularization: 
\begin{equation}
    \mathcal{L}_{d} = \sum_{\mathbf{x}_{\text {surf }}}\left|\boldsymbol{k}_d\left(\mathbf{x}_{\text {surf }}\right)-\boldsymbol{k}_d\left(\mathbf{x}_{\text {surf }}+\epsilon\right)\right|, 
\end{equation}
\begin{equation}
    \mathcal{L}_{orm}= \sum_{\mathbf{x}_{\text {surf }}}\left|\boldsymbol{k}_{o r m}\left(\mathbf{x}_{\text {surf }}\right)-\boldsymbol{k}_{o r m}\left(\mathbf{x}_{\text {surf }}+\epsilon\right)\right|, 
\end{equation}
where $\mathbf{x}_{\text{surf}}$ represents the world coordinates of points on the object's surface. $\boldsymbol{k}_d(\mathbf{x}_{\text {surf }})$ and $\boldsymbol{k}_{o r m}(\mathbf{x}_{\text {surf }})$ is the 
 material of this point. $\epsilon$ is a randomly distributed vector of tiny deformations.
A self-supervised loss is used to regulate the learned diffuse color:
\begin{equation}
    \mathcal{L}_{\text{diff}}=\mathcal{L}_{r g b}\left(\mathbf{C}_{\text{diff}}, \boldsymbol{k}_{\text{diff}} \left(\boldsymbol{x_{\text{surf}}} \right)\right),
\end{equation}
where $\mathbf{C}_{\text{diff}}$ is the diffuse light from the object surface obtained during rendering. 
$\boldsymbol{k}_{\text{diff}}(\mathbf{x}_{\text {surf}})$ is the diffuse light obtained by the MLP.
Overall, the full loss function is:
\begin{equation}
    \mathcal{L} = \mathcal{L}_{rgb} + \omega_1 \mathcal{L}_{d} + \omega_2 \mathcal{L}_{orm} + \omega_3 \mathcal{L}_{\text{diff}},
\end{equation}
where $\omega_1$, $\omega_2$ and $\omega_3$ are three predefined scalars.

\begin{figure}[t]
  \centering
  \includegraphics[width=\linewidth]{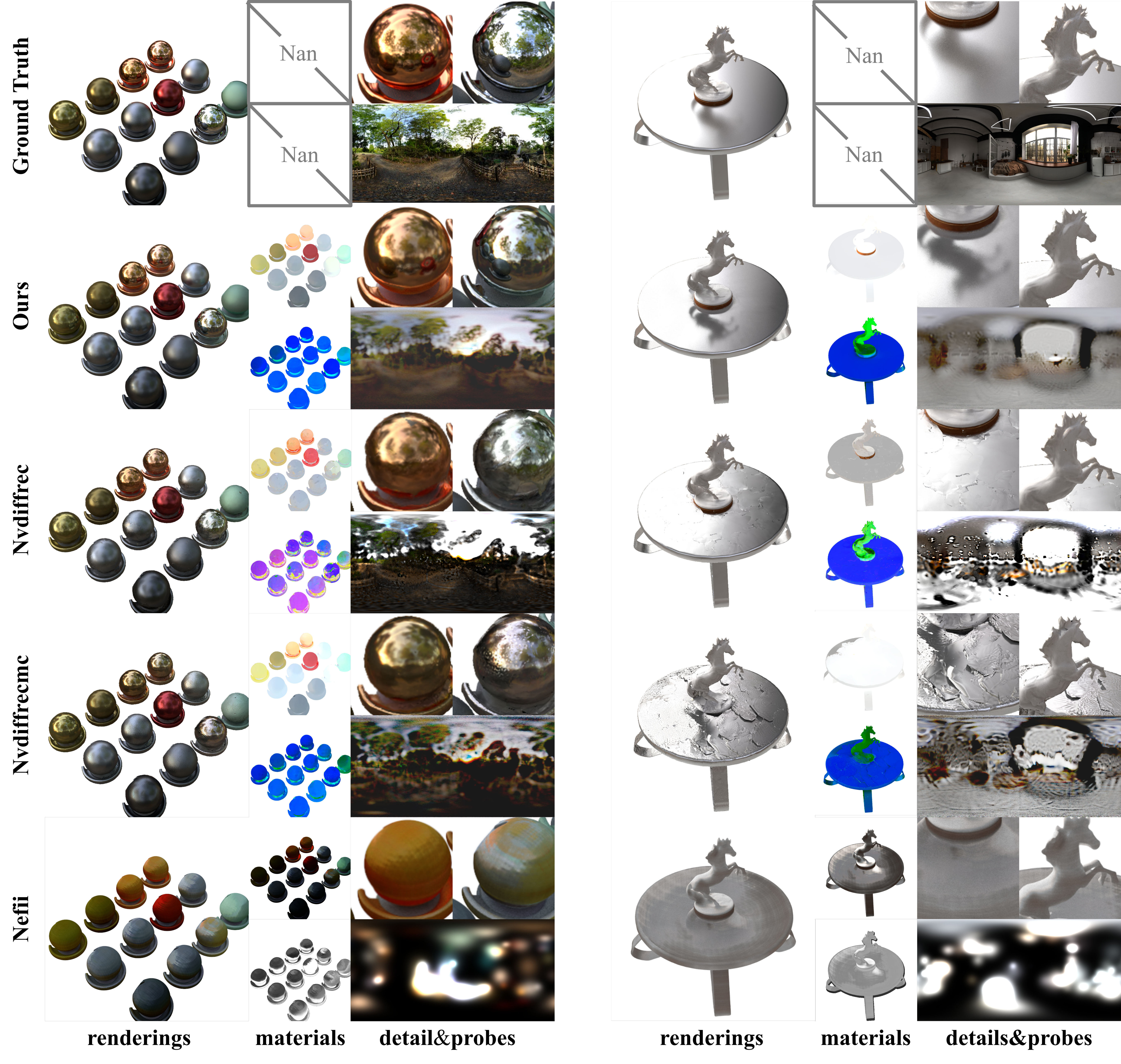}
   \caption{\textbf{Qualitative comparison}. The results of renderings, materials, and environment maps are presented. Note that, the material of Nefii contains only roughness without metalness. Our method achieves the best renderings with clear reflections, compared to other inverse rendering methods. Our method is also superior to others in the disentanglement of materials and environment maps.}
   \label{fig:exp_1}
\end{figure}

\section{Experiments}

\subsection{Implementation Details}
\noindent\textbf{Dataset.} 
We construct a dataset of multiple reflective objects based on the existing single objects to evaluate the performance.
Our dataset consists of 16 groups of object compositions, most of which contain indirect illumination between reflective objects.
We render the composed objects with various environmental lighting in the Blender engine. Each group contains 300 images, with 200 for the training set and 100 for the test set.


\noindent\textbf{Experiment setup.} We optimize the 3D model on 1 RTX 3090 GPU with 24G memory. 
We use the Adam optimizer for the material and the environment map with an initial learning rate of 0.03.
The coefficients of loss function $\omega_1$, $\omega_2$, and $\omega_3$ are set to 0.1, 0.05, and 1, respectively.
The rate of Monte Carlo sampling is commonly set to 128 consistent with the setting in Nvdiffrecmc~\cite{hasselgren2022shape}.


\begin{table}

  \centering
  \caption{\textbf{Quantitative Comparisons}. Ours ($x$) means the $x$ times of sampling in our method. Ours (1) is about 1.5h longer than Nvdiffrecmc~\cite{hasselgren2022shape} as we need additional time to learn the initial geometry.}
  \label{tab:comparison}
  \begin{tabular}{@{}lccccccccc@{}}
    \toprule
    Method & \makecell{NDR \\ \cite{munkberg2022extracting}} & \makecell{NDRMC \\ \cite{hasselgren2022shape}} &\makecell{Nefii \\ \cite{wu2023nefii}} & \textbf{Ours} & \makecell{Ours \\ (w/o Acc.)} & \makecell{Ours \\ (w/o Geo.)} & \makecell{Ours \\ (3)} & \makecell{Ours \\ (1)} \\
    \midrule
    PSNR$\uparrow$ & 26.9 & 25.7 & 22.3 & 28.1 & 28.0 & 24.6 & 28.4 & 27.3\\
    Training time$\downarrow$ & 30min & 45min & 20h & 6.5h & 11.5h & 5h &26.5h & 2.25h \\
    \bottomrule
  \end{tabular}
\end{table}

  

\subsection{Comparison with Baseline}
In this section, we compare our method with Nvdiffrec~\cite{munkberg2022extracting}, Nvdiffrecmc~\cite{hasselgren2022shape}, and Nefii~\cite{wu2023nefii} on our constructed reflective dataset, and the results are shown in~\cref{fig:exp_1}.
Nvdiffrec achieves photo-realistic results in the \textit{metal balls} but fails on the smooth and glossy \textit{table + horse}. Besides, Nvdiffrec cannot well disentangle the materials because it does not consider the shading.
In contrast, Nvdiffrecmc performs well in material learning, while its rendering results are bad. 
These two methods also suffer from indirect illumination from the reflection of inner objects, leading to low-quality environment maps.
Nefii is the recent work that considers the indirect illumination in the radiance field, but it tends to produce low-reflective results and performs badly in these high specular objects.
Compared to these methods, our method can achieve both photo-realistic rendering results and well-disentangled material learning. We can handle highly specular objects, \eg, the metal table. The environment maps learned by our methods are also superior to others.

\begin{figure}[t]
   \centering
   \includegraphics[width=\linewidth]{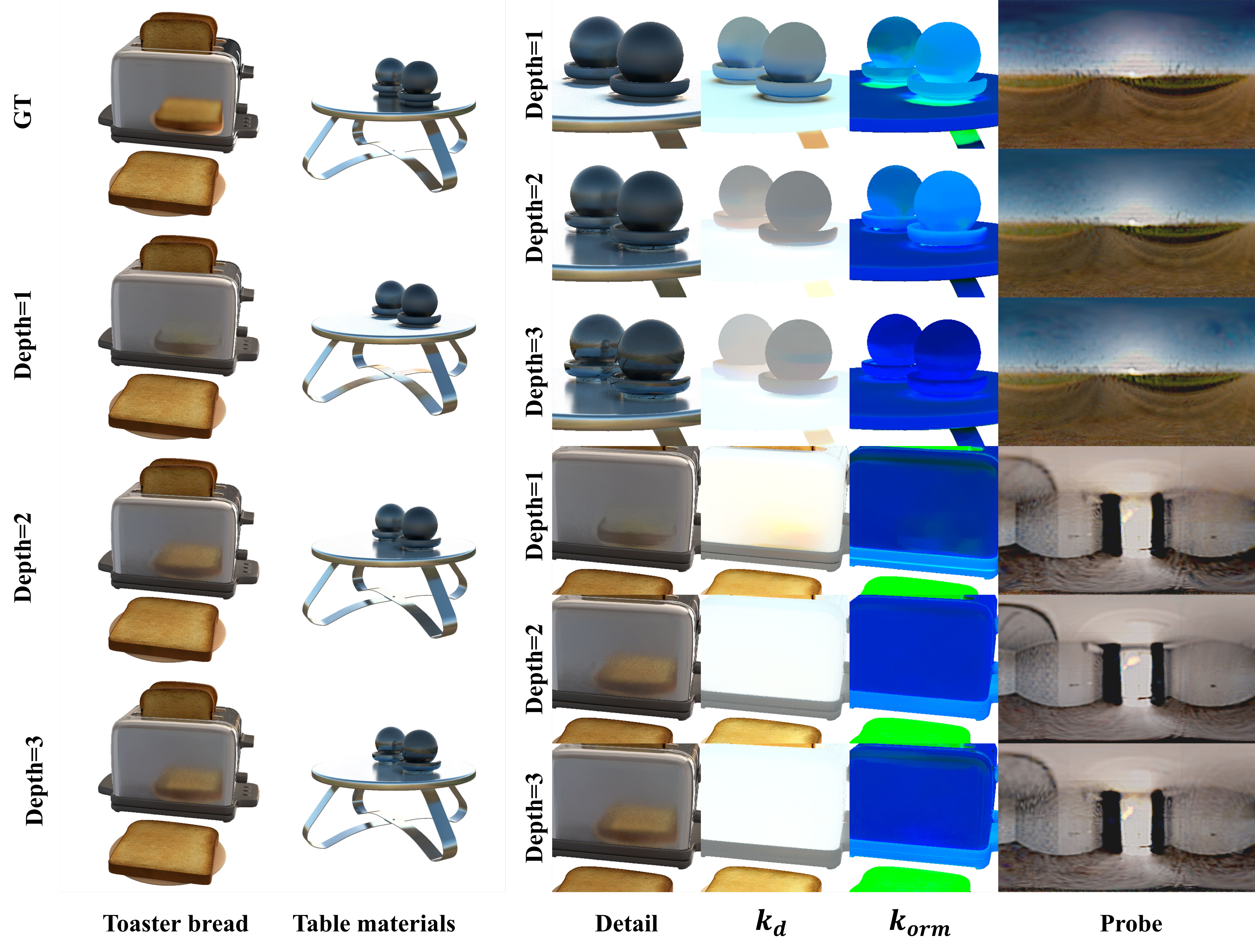}
    \caption{\textbf{Ablation study} on the depth of ray tracing, \ie, the times of Monte Carlo sampling. The results with depth=1 show fewer and darker reflections compared to the ground truth. $k_d$ maps also illustrate the limited capacity to disentangle the material from environmental light, for example, mistaking the diffuse color of the table as the color of the sky. In contrast, with depth=2 or 3, results show more realistic renderings and disentangled materials.}
    \label{fig:exp_2}
\end{figure}

\begin{figure}[!tp]
   \centering
   \includegraphics[width=\linewidth]{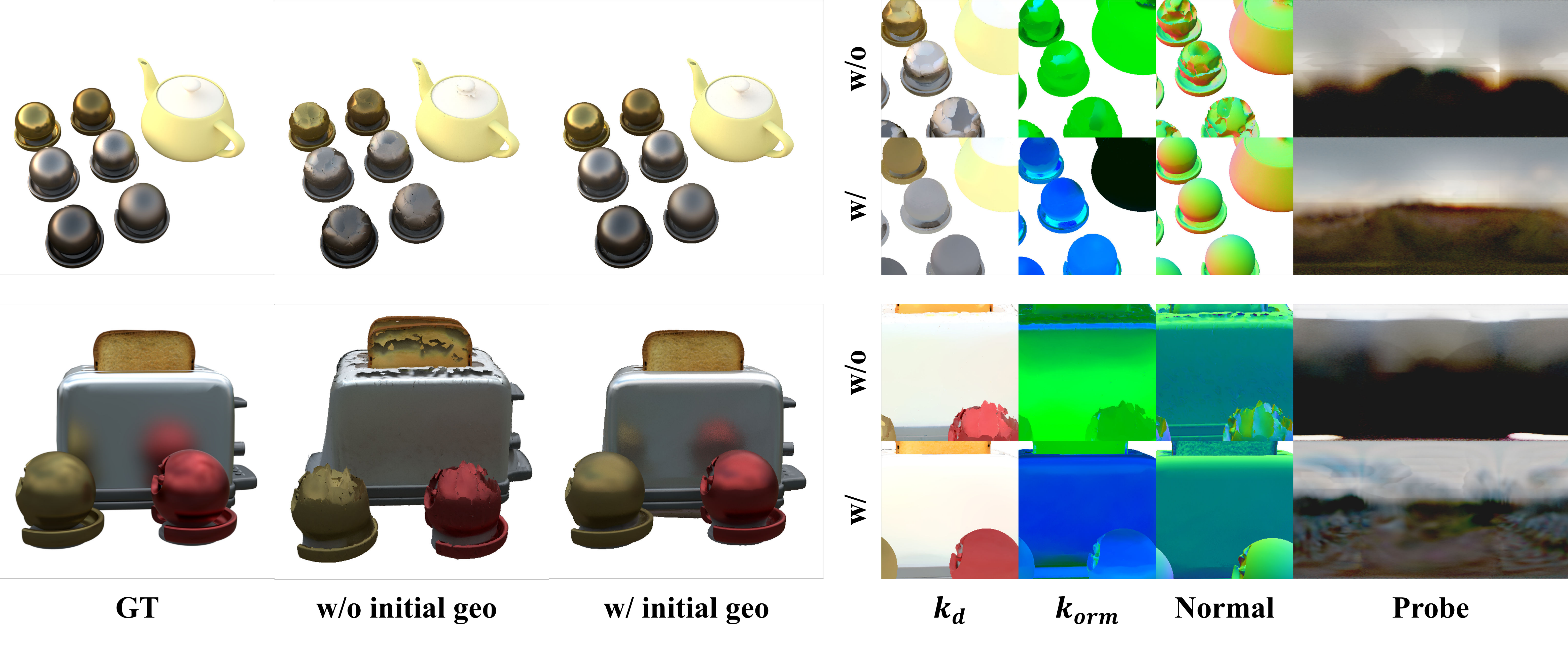}
    \caption{\textbf{Ablation study} on geometric initialization. As shown, a better-quality geometry can significantly improve the material learning and also refine the rendering results.}
    \label{fig:exp_3}
\end{figure}

\subsection{Multi-times Monte Carlo Sampling}
In this section, we conduct an ablation study on the times of Monte Carlo Sampling, \ie the depth of ray tracing. Multi-times sampling considers the indirect illumination, and thus successfully disentangles the environment light and inner reflective light. As shown in~\cref{fig:exp_2}, the environment maps learned by single-time sampling contain noise and shades that are actually the inner reflection. In contrast, multi-times sampling significantly releases the problem, producing a clearer environment map. Single-time sampling also undermines the rendering results, with darker inverted reflections. When the sampling ray is sheltered by objects during path tracing, the ray returns no light, resulting in a dark point. Multi-times sampling returns the reflective light of the sheltering points containing both diffuse and specular light. As a consequence, the rendering results of multi-times sampling show realistic appearance and reflections. 
Besides, it can be seen that results with 2-time sampling are comparable to results with 3-time sampling. \cref{tab:comparison} also quantitatively shows minimal improvement by deeper tracing. However, each additional sampling time induces a 4-times computation increment. Therefore, a 2-times sampling is a more cost-effective setting.
Additionally, we also compare the efficiency between the sampling with and without acceleration. As shown in~\cref{tab:comparison}, our method can make the training speed twice as fast while keeping a comparable PSNR. The degree of acceleration is related to the complexity of geometry, and here we adopt the median in our dataset.

\subsection{Geometry}
\begin{figure}[h]
   \centering
   \includegraphics[width=\linewidth]{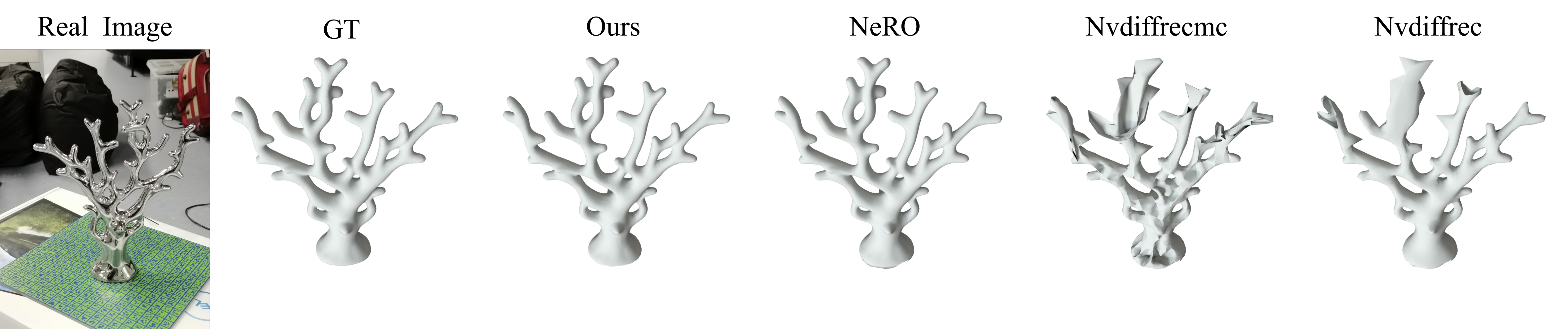}
    \caption{\textbf{Real scene geometry comparison}. We compare our reconstructed geometry with NeRO, Nvdiffrec and Nvdiffrecmc. Our reconstructed geometry demonstrates superior results.}
    \label{fig:exp_geo}
\end{figure}

In this section, we further explore the necessity of the initial geometry discussed in~\cref{sec:geometry} and present our geometry reconstruction results.
As shown in~\cref{fig:exp_3}, without initial geometry, the rendering results show bumpy surfaces with badly learned materials in the area of hollow shape.
Because multi-times sampling amplifies the errors introduced by geometry, the learned materials are even worse than the single-time sampling.
When applying a good-quality initialization, the network bypasses the ambiguity brought by the geometry. It helps our method learn well-disentangled materials and finally produces high-quality rendering results.
Quantitative comparison in~\cref{tab:comparison} also demonstrate the necessity of geometry initialization. In addition, we showcase a real-scene geometry reconstruction result of the NeRO~\cite{liu2023nero} dataset in~\cref{fig:exp_geo}. We further use Chamfer Distance to quantitatively evaluate our geometric quality and present the results in~\cref{tab:geometry}.

\begin{table}[h]

  \centering
  \caption{\textbf{Geometry qualitative comparison}.We use the Chamfer Distance ($\downarrow$) to evaluate our reconstructed geometry. As shown in the table, we obtain the best results in multiple scenes.}
  \label{tab:geometry}
  \begin{tabular}{ccccc}
    \toprule
    Dataset & Nvdiffrec~\cite{munkberg2022extracting} & Nvdiffrecmc~\cite{hasselgren2022shape} &NeRO~\cite{liu2023nero} & Ours \\
    \midrule
    Materials & 0.016 & 0.016 & 0.0057 & \textbf{0.0030} \\
    Coral & 0.28 & 0.25 & 0.13 & \textbf{0.13}  \\
    \bottomrule
  \end{tabular}
\end{table}

\begin{figure}[t]
   \centering
   \includegraphics[width=\linewidth]{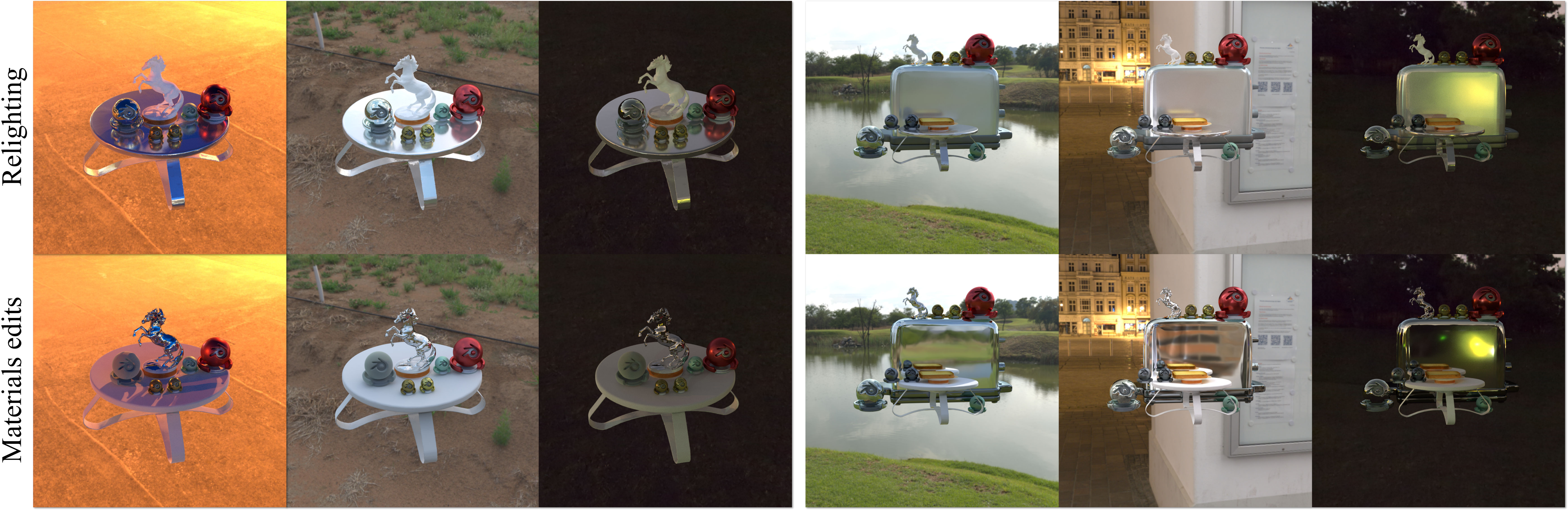}
    \caption{\textbf{Relighting and editing}. We compose the reconstructed objects in a unified scene and change the environmental light. The relighting results show our strong ability to disentangle the light and shading. We also perform material edits which shows the flexibility in wide applications.}
    \label{fig:relighting}
\end{figure}

\begin{figure}[h]
   \centering
   \includegraphics[width=\linewidth]{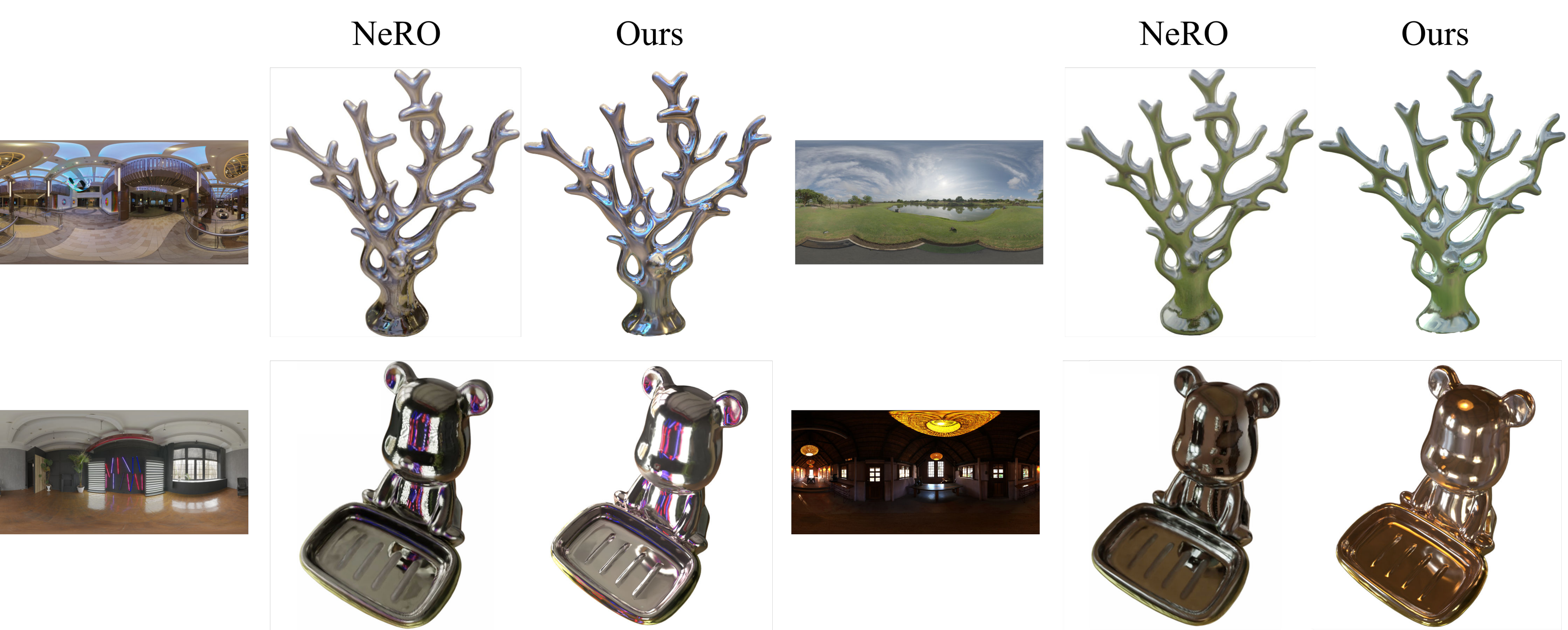}
    \caption{\textbf{Relighting results of real scenes}. Our method demonstrates more realistic results under several lighting conditions than NeRO.}
    \label{fig:real}
\end{figure}

\subsection{Relighting and Material Editing}
The disentangled environment map and material empower our methods to relight and edit reconstructed objects in the downstream application. 
Our method can also be seamlessly plugged into the industrial pipeline.
As shown in~\cref{fig:relighting}, our method enables the arbitrary combination of reconstructed objects. We can easily edit their metalness, roughness, and albedo color by manipulating the learned materials. Due to the well-disentangled shading, our rendering results are natural and realistic in all five environment maps, even the point light in a dark environment. They also perform well after editing materials thanks to the well-learned material map. For example, after we increase the metalness and reduce the roughness of the teapot, the teapot clearly reflects the neighboring objects on its surface. In another case, where we extremely increase the reflectance and change the base color of the toaster, it accomplishes to reflect the scene and other objects. To showcase further applications of our method, we conduct experiments on real-world datasets captured by NeRO~\cite{liu2023nero} and present the comparison results in~\cref{fig:real}. It can be seen that Ref-MC$^2$ achieves more realistic results under different lighting conditions.

\section{Conclusion and Limitations}

In conclusion, our Ref-MC$^2$ introduces multi-times Monte Carlo sampling into the inverse rendering pipeline to model the indirect illumination.
It improves the performance in scenes with complex inter-reflections.
However, increment of sampling times significantly increases computational consumption and makes the pipeline highly geometry-sensitive.
To solve the challenge of computational efficiency, based on the Lambert model, 
we simplify the BRDF for indirect lighting, which allows us to reduce the number of ray traces.
To improve the geometry quality, we adopt SDF-based architecture to get an initial geometry and refine a design with Spherical Gaussian encoding for reflective objects. We further use Flexicubes to take the initial mesh into the differentiable rendering pipeline that learns disentangled materials. 
Our Ref-MC$^2$ still has several limitations.
The major limitation is that the 2-time sampling cannot handle extremely reflective objects, \eg, mirrors, as the specular energy hardly degrades after reflections. Besides, the training time needs to further reduce in future time.

\section{Acknowledgements}
This work was supported by NSFC (62201342) and  Shanghai Science and Technology Major Project (2021SHZDZX0102). We also thank Student Innovation Center of SJTU for providing GPUs.

\clearpage
{\small
\bibliographystyle{ieee_fullname}
\bibliography{ref}
}

\end{document}